\documentclass[letterpaper, 10 pt, conference]{ieeeconf}

\IEEEoverridecommandlockouts                              % This command is only needed if 
\usepackage[pdftex]{graphicx}
\usepackage{url}
\usepackage{amsmath,amssymb}
\usepackage{bm}
\usepackage{color, colortbl}
\usepackage[colorlinks,bookmarksopen,bookmarksnumbered,citecolor=red,urlcolor=red]{hyperref}

\hypersetup
{
	pdftitle = {Whole-body control for quadrupedal locomotion on challenging terrain},
	pdfauthor = {Shamel Fahmi},
	pdfsubject = {RA-L manuscript},
	pdfkeywords = {whole-body control, legged robots, challenging locomotion, and
	terrain mapping},
	pdftoolbar = true,
	colorlinks = true,
	linkcolor = black,
	citecolor = black,
	urlcolor = black,
}

\usepackage[usenames,dvipsnames]{xcolor}%\usepackage{xcolor,colortbl}
\definecolor{blue_iit}{RGB}{51,51,255}
\usepackage{algpseudocode}
\usepackage{algorithm}
\usepackage{glossaries}
\usepackage[tight]{units}
\usepackage[normalem]{ulem} % to strike out text, use: \sout{text}
\usepackage{cancel}
\definecolor{Gray}{gray}{0.9}

\newacronym{hyq}{HyQ}{Hydraulically actuated Quadruped}

\newacronym{lf}{LF}{Left-Front}
\newacronym{rf}{RF}{Right-Front}
\newacronym{lh}{LH}{Left-Hind}
\newacronym{rh}{RH}{Right-Hind}

\newacronym{haa}{HAA}{Hip Adduction-Abduction}
\newacronym{hfe}{HFE}{Hip Flexion-Extension}
\newacronym{kfe}{KFE}{Knee Flexion-Extension}

\newacronym{imu}{IMU}{Inertial Measurement Unit}
\newacronym{dofs}{DoFs}{Degrees of Freedom}
\newacronym{rt}{RT}{Real Time}

\newacronym{com}{CoM}{Center of Mass}
\newacronym{cop}{CoP}{Center of Pressure}
\newacronym{zmp}{ZMP}{Zero Moment Point}
\newacronym{icp}{ICP}{Instantaneous Capture Point}
\newacronym{cmp}{CMP}{Centroidal Moment Pivot}
\newacronym{grfs}{GRFs}{Ground Reaction Forces}

\newacronym{ls}{LS}{Least Square}

\newacronym{slip}{SLIP}{Spring Loaded Inverted Pendulum}
\newacronym{eom}{EoM}{Equation of Motions}
\newacronym{qp}{QP}{Quadratic Program}
\newacronym{sqp}{SQP}{Sequential Quadratic Programming}
\newacronym{mic}{MIC}{Mixed-Integer Convex}
\newacronym{cmaes}{CMA-ES}{Covariance Matrix Adaptation Evolution Strategy}
\newacronym{ara}{ARA*}{Anytime Repairing A*}
\newacronym{pca}{PCA}{Principal Component Analysis}
\newacronym{cpg}{CPG}{Central Pattern Generator}
\newacronym{wbc}{WBC}{Whole-Body Control}

\newacronym{pd}{PD}{Proportional-Derivative}

\newacronym{mpc}{MPC}{Model Predictive Control}
\newacronym{nmpc}{NMPC}{Nonlinear Model Predictive Control}
\newacronym{awbc}{c$^3$WBC}{Compliant Contact Consistent Whole-Body Control}
\newacronym{swbc}{sWBC}{Standard Whole-Body Control}
\newacronym{c3wbc}{c$^3$WBC}{Compliant Contact Consistent Whole-Body Control}
\newacronym{ste}{TCE}{Terrain Compliance Estimator}
\newacronym{c3}{\texttt{c}$^3$}{compliant contact consistent}

\newacronym{stance}{STANCE}{\textbf{S}oft \textbf{T}errain \textbf{A}daptation a\textbf{n}d \textbf{C}ompliance \textbf{E}stimation}

\newacronym{wbopt}{WBOpt}{Whole Body Optimization}

\newacronym{hc}{HC}{Hunt and Crossley's}
\newacronym{kv}{KV}{Kelvin-Voigt's}

\newacronym{wllsr}{WLLSR}{Weighted Linear Least Squared Regression}

\newacronym{mae}{MAE}{Mean Absolute Tracking Error}

\newacronym{ode}{ODE}{Open Dynamics Engine}
\newacronym{lip}{LIP}{Linear Inverted Pendulum}
\newacronym{srbd}{SRBD}{Single Rigid Bidy Dynamics}

\newcommand{\Rnum}{\mathbb{R}} % Symbol fo the real numbers set

\newcommand{\vect}[1]{\mathbf{#1}} %vector bold

\newcommand{\mat}[1]{\ensuremath{\begin{bmatrix}#1\end{bmatrix}}}	% matrix
\newcommand{\atandue}{\textrm{atan2}}

\newcommand\BibTeX{{\rmfamily B\kern-.05em \textsc{i\kern-.025em b}\kern-.08em
T\kern-.1667em\lower.7ex\hbox{E}\kern-.125emX}}

\definecolor{LightBlue}{RGB}{0.4,0.4,1}

\title{\LARGE \bf CLIO: a Novel Robotic Solution for Exploration and Rescue Missions in Hostile Mountain Environments}
\author{Michele Focchi$^{1}$, Mohamed Bensaadallah$^{2}$, Marco Frego$^{3}$, Angelika Peer$^{3}$, \\Daniele Fontanelli$^{4}$, Andrea Del Prete$^{4}$, Luigi Palopoli$^{1}$\vspace{-1.0cm}
\thanks{$^1$ The authors are with the Dipartimento di Ingegneria and Scienza dell'Informazione (DISI), University of Trento. Email:  \href{mailto:name.surname@unitn.it}{name.surname@unitn.it}}
\thanks{$^2$ The author is with the Department of Electronics, University of Batna 2, Algeria. Email: \href{mailto:m.bensaadallah@univ-batna2.dz}{m.bensaadallah@univ-batna2.dz}}
\thanks{$^3$ The authors are with the Faculty of Science and Technology, Free University of Bozen-Bolzano. Email:  \href{mailto:name.surname@unibz.it}{name.surname@unibz.it}}
\thanks{$^4$ The authors are with Dipartimento di Ingegneria Industriale (DII), University of Trento Email:  \href{mailto:name.surname@unitn.it}{name.surname@unitn.it}. The publication was created with the co-financing of the European Union FSE-REACT-EU, PON Research and Innovation 2014-2020 DM1062 / 2021.} }

\begin{document}
\maketitle
\thispagestyle{empty}
\pagestyle{empty}

\begin{abstract}%150-250 word abstract
  Rescue missions in mountain environments are hardly achievable by
  standard legged robots---because of the high slopes---or by flying
  robots---because of limited payload capacity.  We present a 
  concept for a rope-aided climbing robot which can negotiate
  up-to-vertical slopes and carry heavy payloads.  The robot is
  attached to the mountain through a rope, and it is equipped with a leg
  to push against the mountain and initiate jumping maneuvers.
  Between jumps, a hoist is used to wind/unwind the rope to move
  vertically and affect the lateral motion.  This simple (yet
  effective) two-fold actuation allows the system to achieve high
  safety and energy efficiency. Indeed, the rope prevents the robot
  from falling while compensating for most of its weight, drastically
  reducing the effort required by the leg actuator. We also present an
  optimal control strategy to generate point-to-point trajectories
  overcoming an obstacle.  We achieve fast computation time ($<$1 s)
  thanks to the use of a custom simplified robot model.  We validated
  the generated optimal movements in Gazebo simulations with a
  complete robot model with a $<5\%$ error on a 16 $m$ long jump, 
  showing the effectiveness of the proposed
  approach, and confirming the interest of our concept.  Finally, we
  performed a reachability analysis showing that the region of
  achievable targets is strongly affected by the friction properties
  of the foot-wall contact.
\end{abstract}

\section{Introduction}\label{sec:introduction}

% Strong background references addresses "what is already known" 
A climbing robot is a robot that moves on a vertical surface.  The
idea was first suggested in a seminal work by Nishi et
al.\cite{Nishi1986} and has been revisited throughout for different
application domains, including
%% The idea of developing climbing robots was expected for years but only in 1986, 
%% Nishi et al. presented in their seminal paper \cite{Nishi1986} 
%% the design of a robot that can walk on a vertical surface. During the last thirty years, many 
%% prototypes of climbing robots have been proposed for specific applications to replace 
%% humans in dangerous and difficult tasks, including inspection of
nuclear power 
plants \cite{Briones1994}, high-rise building cleaning \cite{Nansai2016}, bridge 
maintenance \cite{Wang2016}, search and rescue missions \cite{Eich2008}. 
However, % Mention a Gap in knowledge
there are still ways to climb walls not yet studied or reported \cite{bio_inspired2015}. 

% Introduction of the problem 
The research work done so far on climbing robots is reviewed and
classified in \cite{Fang2022} and \cite{A.Hajeeretall2020},
respectively, according to the adhesion technique (electrostatic, magnetic, pneumatic, etc.) \cite{longo2008}
used to attach the robot to the wall surface, and locomotion mechanism (i.e. wheel-Driven, tracked, legged, etc.)
\cite{Chu2010}, utilized for climbing.  The attachment and motion
mechanisms are basically considered as the main problems when
designing climbing robots. Additional requirements of equal importance
come from the application and include (R1) safety, (R2) quickness in
emergency situations, (R3) carrying payloads, and (R4) avoiding
obstacles \cite{Schmidt2013}.
f
% Summarizing what's coming in the introduction
%In the coming two paragraphs, we give the reasons behind using a rope/wire 
%as an attachment method. After that, we suggest the jumping as moving technique.

\begin{figure}[t]
	\centering
	\includegraphics[width=0.45\columnwidth]{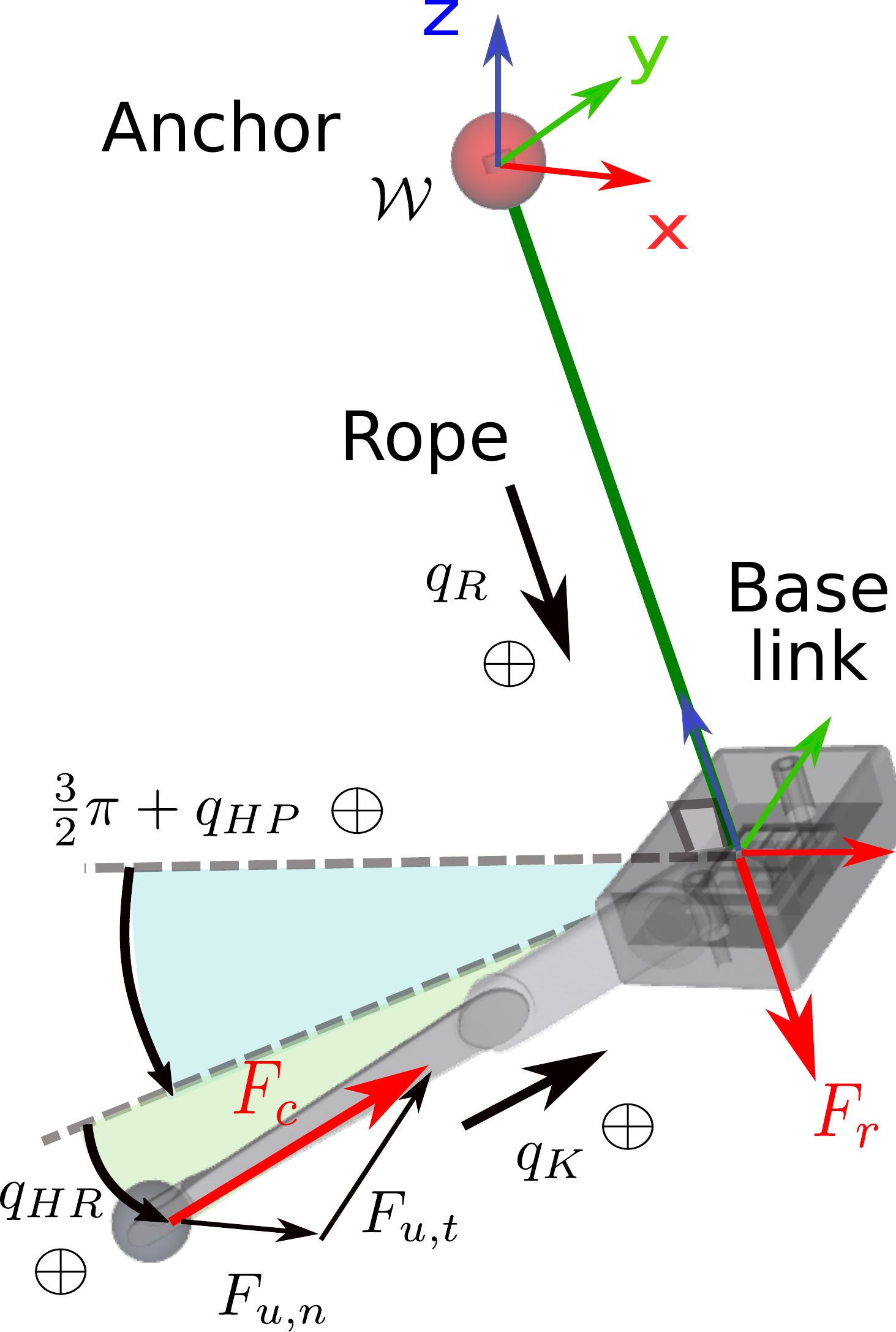}
		\includegraphics[width=0.52\columnwidth]{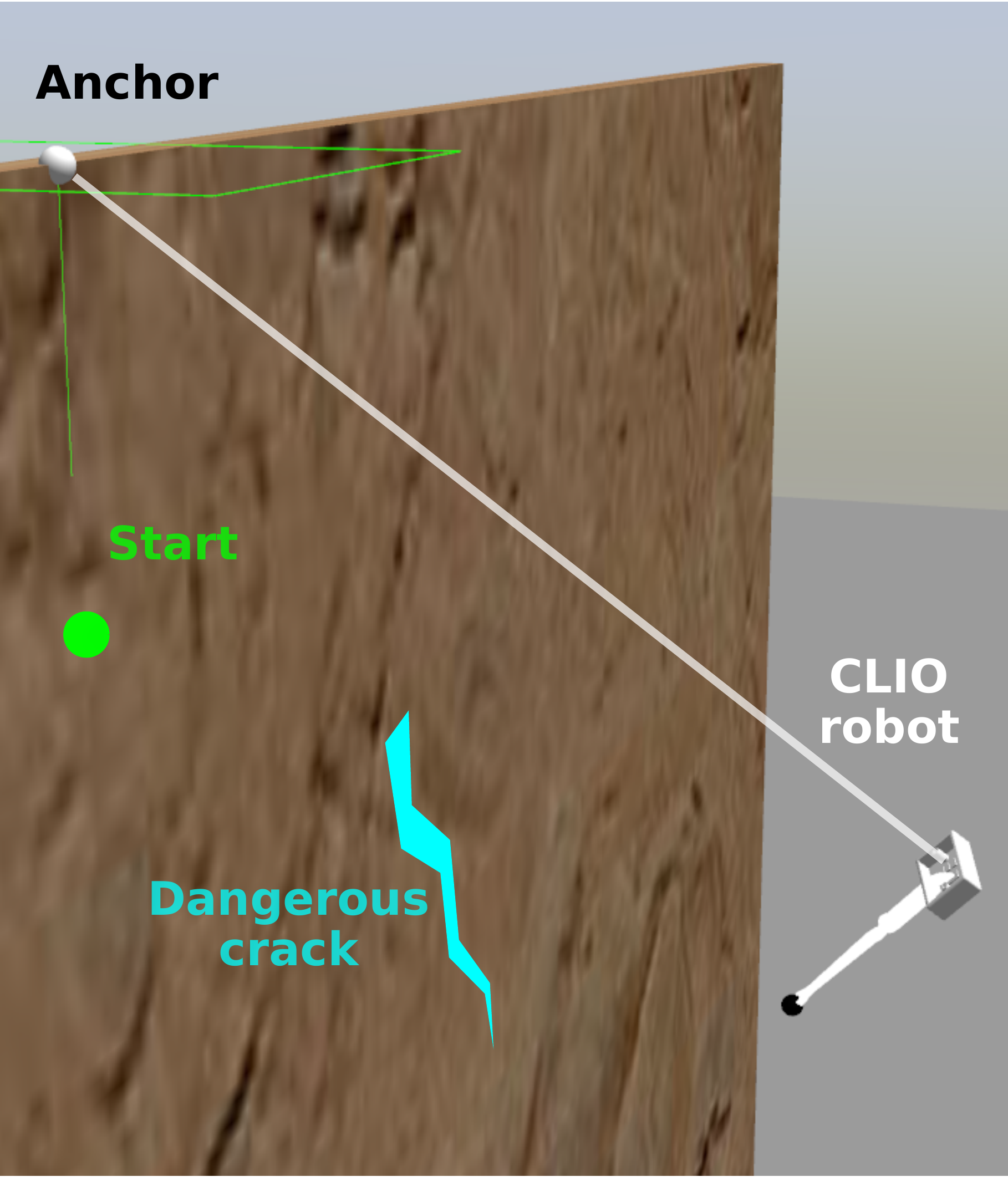}
	\caption{\small (left) Kinematic model of the CLIO robot with  standard definitions. 
		The anchor frame is aligned with an inertial ($\mathcal{W}$) frame. (right) Realistic Gazebo simulation of a jump.}
	\label{fig:3dmodel}
\end{figure}

%% Solution for Attachment technique : USING A ROPE + main advantages   
Human climbers inspired the use of a rope mainly for safety measures (R1)
to support the weight of people during facade cleaning, firefighting and rescue missions
\cite{Nansai2016}. 
In this paper, we combine the use of a rope with a leg, which can be retracted and extended
very quickly. This allows the robot to jump away from the mountain wall, while the rope
can be used to control its motion.

%Legged robots---as well as mobile robots---cannot walk on slopes with too high inclinations \cite{Abdalla12020} because limited friction results in slippage and falling.
%Instead, using a rope introduces an external force to compensate 
%for gravity (along the rope direction) and 
%allows the contact forces to satisfy the friction constraints (i.e., laying in 
%the middle of the friction cones) which ensures inherent safety (R1).
%
%% second advantage of using a rope : increase speed of movment 
Additionally, as shown by Wang et al. \cite{Wang2014} on dragline 
locomotion bio-inspired from spiders, the aid of rope can dramatically increase the locomotion speed (R2) 
by a winding and releasing mechanism, therefore being a preferable solution in applications
that require a prompt intervention, such as Search $\&$ Rescue missions.  
%  
%% Third advantage 
Furthermore, ropes can support much heavier payloads (R3) than robots relying on sticky/adhesive approaches~\cite{Kim2008,Riskin2009}
because of the limited tangential component of the adhesive and leg actuation.
%the amount of payload that can be carried by sticky/adhesive based climbing 
%robots \cite{Kim2008,Riskin2009} is very limited due to the small tangential component of 
%the adhesive and leg actuation. Conversely, a feature of rope to be ideally 
%inextensible\footnote{The maximum payload will be limited by the maximum force deliverable 
%by the winding mechanism, e.g. a hoist, which can be extended using a gearbox.} \ADP{I don't see the connection between the rope being inextensible and the payload. The payload is large because gravity is compensated through the rope force instead of the contact force with the wall.}
%will allow the robot to carry a heavier payload (R3). 
%Difficulties due to the dynamics when using a rope
However, having the robot attached to a rope poses some challenges.
First, the robot is under-actuated because it cannot \textit{fully} control the position of its center of mass when 
not in contact with the wall. Second, the rope represents a \textit{unilateral} 
constraint (i.e. it can only pull and not push), which further complicates the already hybrid 
dynamics and the low control authority of this class of robots. 
% Providing a solution for second problem: locomotion technique, we
% can reach the target with jumps
Finally, rather than slowly taking steps as a walking mechanism, the robot can take one or multiple 
jumps like the rapid jumping Salto-1P shown in \cite{Haldane2017} to reach the desired locations, 
and overcoming obstacles. Some early work on planning jumping trajectories for a rope-aided  
robot has been performed in \cite{mingo21}, but no physical validation was reported.

%Difficulties in the jumping technique
To summarize, the key features of jumping with a rope %that can be
% released
are related to the locomotion speed and the ability to address up to
\textit{vertical} inclinations.  Moreover, the resulting motion of the
robot (and so the possibility to reach the target) depends on: 1) the
impulse exerted on the wall at lift-off, 2) the winding/unwinding of
the rope.
%
%Optimization for plannig 
Therefore, a planning strategy for this kind of motions should take 
into account both these factors, besides the under-actuation, 
the constraints posed by the rope, the actuator limits and the 
contact interaction (i.e. friction). 

To tackle these issues, numerical optimization comes as an attractive
solution for these planning problems \cite{Nguyen2019,Ding2020}, since
it allows to minimize some optimality criteria while ensuring that
the physical constraints are satisfied along the planning horizon.
%
%Indeed, casting locomotion planning as an optimization problem
%allows one to represent high-level tasks and system dynamics using cost functions and 
%constraints. 
In this framework, different goals can also be pursued, such as minimizing energy 
consumption or reaching the target in minimum time. 

In this work, we present a robotic platform called CLIO
(CLImbing rObot) that is able to reach desired targets on a vertical
(or slanted) wall with different frictional properties.  We also
propose a planning approach based on numerical optimization, to solve
the jumping problem employing a simplified model of the dynamics.
Hence, the contributions of the paper can be summarised in what
follows:
\begin{itemize}

\item A conceptual design of a jumping robot platform CLIO;
\item An optimal control formulation to generate a jump motion to
  reach a desired target while overcoming an obstacle, based on a
  simplified model of the system, which results in reduced computation
  time;
\item Simulation experiments to validate the effectiveness of the
  proposed approach, both using the simplified model and in a more
  realistic (Gazebo) simulation, considering the full dynamics of the
  robot.
\end{itemize}
%
%\subsection{Outline}
%
As an additional result, we propose a reachability analysis that shows
that the region of achievable targets is limited by the friction
coefficient at the lift-off location.

The paper is organized as follows: Section \ref{sec:robot} gives an
overview of the robot platform and derives a simplified model of it.
Section \ref{sec:motionP} describes the optimization problem to plan
jump trajectories based on the simplified model.  Simulation results
are presented in Section \ref{sec:results}.  Finally, we draw the
conclusions in Section \ref{sec:conclusion}.

\section{Robot Description}\label{sec:robot}
\subsection{Full robot description}
This section presents the kinematic model of the CLIO robot (see Fig. \ref{fig:3dmodel}(left)). 
We model the robot as a serial kinematic chain with $n=9$ \gls{dofs} represented by the configuration vector $\vect{q} \in \Rnum^n$. 
We model the attachment between the anchor point (a fixed link) and the rope with 2 \textit{passive} (rotational) joints.
The rope is modeled as an \textit{actuated} prismatic joint ($q_{R}$), 
followed by 3 \textit{passive} rotational joints to model 
the connection between the rope and the \textit{base link} of the robot\footnote{It would be equivalent to allocate 3 joints at the anchor point and 2 at the base. 
To avoid redundancy in the representation, it is necessary  to have only one passive joint along the rope axis.}. 

The propulsion mechanism that allows the robot to jump is a 3-DoFs
robotic leg with a point-like foot that enables to exert a \textit{pure} Cartesian force (no moment) on the wall. 
The leg has two \textit{subsequent} rotational joints, called  hip pitch ($q_{HP}$ about the Y axis (green)) and hip 
roll ($q_{HR}$ about the X axis (red)). These joints are useful to align the leg to the thrusting impulse. 
This enables to avoid \textit{centroidal} moments that
 would pivot the robot around the rope axis.
% dealing with this is out of the scope of this paper (and part of future works). 
Finally, a prismatic knee ($q_{K}$) joint is used to generate the \textit{thrusting} impulse.
With the actual design, the robot will not be able to stabilize itself on the wall.
However, the design of a landing and stabilization mechanism is out of the scope of this work,
where we focus on the jump. 

%\begin{figure}
%	\centering
%	\includegraphics[width=0.7\columnwidth]{figs/3dmodel.pdf}
%	\caption{\small Kinematic model of the CLIO robot with  standard definitions. The anchor frame is aligned with an inertial ($\mathcal{W}$) frame.}
%	\label{fig:3dmodel}
%\end{figure}

The dynamics equation can be written as:  
\vspace{-0.2cm}
\begin{align}
\vect{M} (\vect{q}) \vect{\ddot{q}} + \vect{h}(\vect{q},\vect{\dot{q}}) = \mat{\vect{0}_{5 \times 1} \\ \boldsymbol{\tau}_a} + \vect{J}^T \vect{F}_c 
\end{align}

where $\vect{M} \in \Rnum^{n \times n}$ is the inertia matrix; $\vect{h} \in \Rnum^n$ represents the bias terms (Centrifugal, Coriolis and Gravity); and 
$\vect{J} \in \Rnum^{3 \times n} $ is the Jacobian relative to the contact point mapping the contact force $\vect{F}_c \in \Rnum^3$. 
Unless specified, all  vectors are expressed in an inertial $\mathcal{W}$ frame (attached to the anchor). 
The under-actuation is captured by the vector $\vect{0}_{5 \times 1}$, while 
the efforts of the actuated joints are grouped into the vector $\boldsymbol{\tau}_a = \mat{ \boldsymbol{\tau}_R & \boldsymbol{\tau}_{\text{leg}}}^T \in \Rnum^4$.

\vspace{-0.2cm}
\subsection{Simplified Model}
\begin{figure}[tbp]
\centering
\includegraphics[width=0.6\columnwidth]{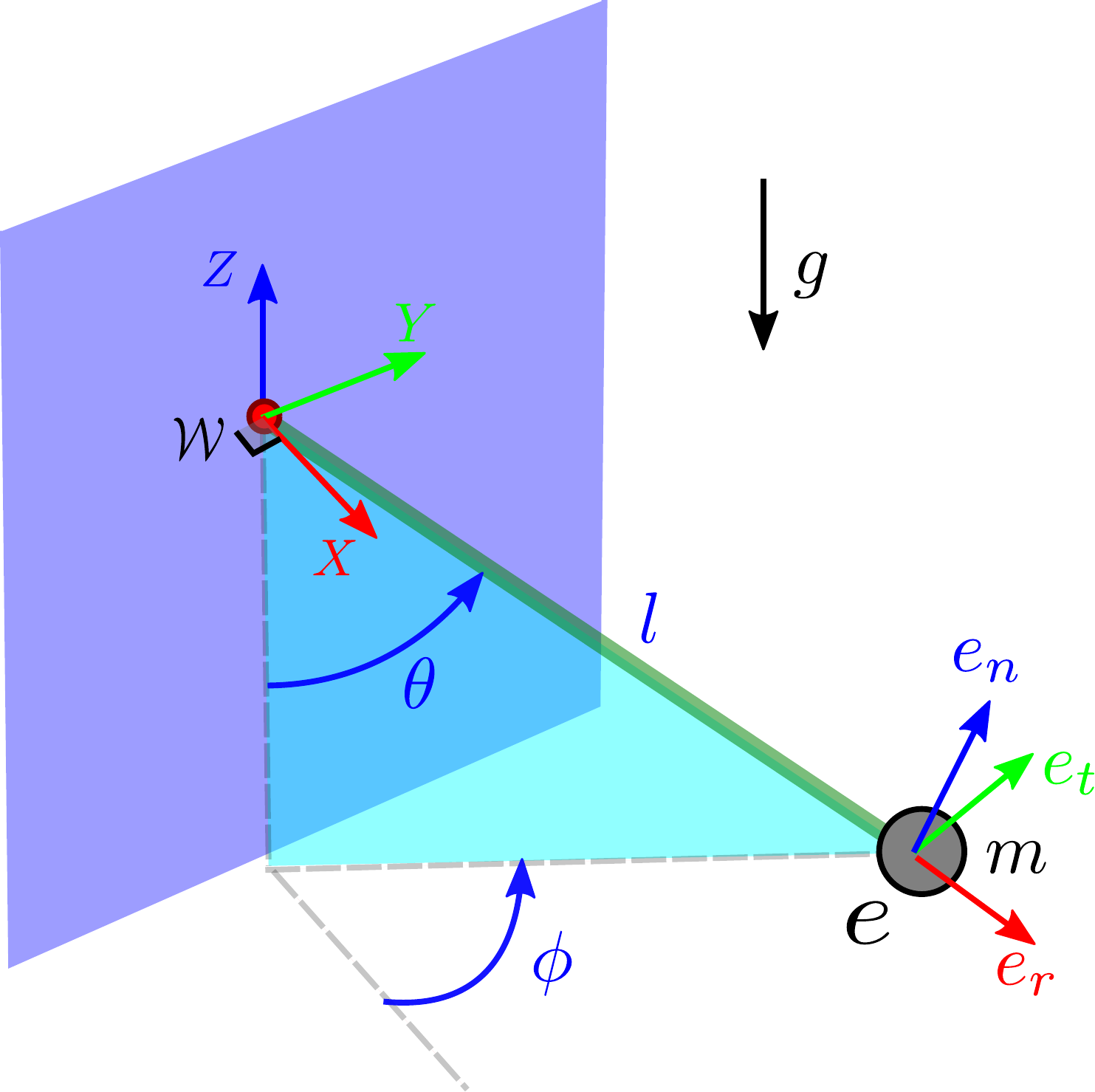}
\caption{\small Logical scheme of the simplified model}
\vspace{-0.3cm}
\label{fig:sp}
\end{figure}

Dealing with the full dynamics of the robot can be time consuming, due to the 
high number of states and constraints involved. Therefore, in this section, we derive a lower dimensional model 
(3 \gls{dofs}) that captures the dominant dynamics of the system by making the following assumptions: 1) the mass is entirely concentrated in the body attached to the rope, 
2) we can unwind and rewind the rope that is assumed to be inextensible, rigid, and remains completely elongated (i.e., it cannot bend), 
3) the rewinder mechanism can pull the rope when winding, but it cannot push (i.e., it can only act as a brake during the unwinding phase).
A geometric sketch of the system is shown in Fig. \ref{fig:sp}, where the pitch angle is $\theta$, the yaw angle is $\phi$, 
and $l$ is the length of the rope. 

Our input variables to the system are: 1) the rewinder pull action or
braking action $\mathbf{F}_r$ oriented along the rope, 2) an impulsive
pushing force $\mathbf{F}_u$ that the robot can generate when it is
attached to the mountain wall. 

Let $e$ be a frame attached to the point mass $m$, and let $\mathcal{W}$ be the world
frame attached to the anchor point of the pendulum.  If we
consider the homogeneous transformation from $e$ to $\mathcal{W}$, we can write: 
	\[
\vect{T}_e^\mathcal{W} = \begin{bmatrix}\vect{R}_z(\phi) & \mathbf{0}_{3 \times 1}  \\ \mathbf{0}_{1 \times 3}  & 1\end{bmatrix}  \begin{bmatrix} \vect{R}_y( \frac{\pi}{2} - \theta) & \mathbf{0}_{3 \times 1}   \\ \mathbf{0}_{1 \times 3}  & 1\end{bmatrix}  
\begin{bmatrix} \vect{I}_{3 \times3} &    \begin{matrix} l \\0 \\0 \end{matrix}    \\ \mathbf{0}_{1 \times 3}  & 1\end{bmatrix}  
	\]
Simplifying $\vect{T}_e^\mathcal{W}$, gives 
	\[
	\vect{T}_e^\mathcal{W} = \begin{bmatrix} c_\phi s_\theta & -s_\phi & c_\phi c_\theta & l c_\phi s_\theta\\
		s_\phi s_\theta & c_\phi & c_\theta s_\phi & l s_\phi s_\theta\\
		-c_\theta & 0 & s_\theta &-l c_\theta\\
		0 & 0 &0 &1\end{bmatrix} 
	\]		
where $c_x$ is a shorthand for $\cos x$ , and $s_x$ is a shorthand for $\sin x$.
The position $\mathbf{p}$ of the mass  can be extracted from the transformation matrix $\vect{T}_\mathcal{B}^\mathcal{W}$ as: 
\begin{align}  
		\mathbf{p} &= \begin{bmatrix} X\\ Y\\ Z \end{bmatrix} =
		\begin{bmatrix}   
            l s_\theta c_\phi\\
			l s_\theta s_\phi\\
			-l c_\theta \end{bmatrix} \label{eq1}						
\end{align}

while the axes $\vect{e}_r$, $\vect{e}_t$, $\vect{e}_n$ are the columns of 	the top-left sub-matrix of $\vect{T}_e^\mathcal{W}$, respectively.
%\begin{align}
%	\mathbf{e}_r &= \mat{	c_\phi s_\theta\\
%		s_\phi s_\theta\\
%		-c_\theta} \quad 
%	\mathbf{e}_t = \mat{-s_\phi\\
%		c_\phi\\
%		0} \quad 
%	\mathbf{e}_n = \mat{c_\phi c_\theta \\
%		s_\phi c_\theta\\
%		s_\theta}		
%\end{align}
%
%From equation \ref{eq1}, we can compute the velocities and the kinetic energy:   
From equation \eqref{eq1}, the velocities along the Cartesian axes are:
\begin{align*}
	\dot{X} &= 
	l c_\theta c_\phi \dot{\theta} - l s_\theta s_\phi \dot{\phi} + s_\theta c_\phi \dot{l} \\
	\dot{Y} &= l c_\theta s_\phi \dot{\theta} + l s_\theta c_\phi \dot{\phi} + s_\theta s_\phi \dot{l} \\
	\dot{Z} &= l s_\theta  \dot{\theta} - c_\theta \dot{l} 
\end{align*}
Next, the velocity of the mass squared:
\begin{align*}
	v^2 &= \dot{X}^2 + \dot{Y}^2 + \dot{Z}^2 =  l^2 \dot{\theta}^2 + l^2 s_\theta^2 \dot{\phi}^2 + \dot{l}^2 		
%	&=  l^2 \dot{\theta}^2 + l^2 s_\theta^2 \dot{\phi}^2 + \dot{l}^2\\
\end{align*}
Thus, the kinetic energy $K$ and potential energy $V$:  
\begin{align*}
	K &=\frac{ m }{2} v^2 =  \frac{m}{2} l^2 \left(\dot{\theta}^2 + s_\theta^2 \dot{\phi}^2 \right) + \frac{m }{2} \dot{l}^2  \\
    V &= mgz = -mgl c_\theta   \label{eq3}
\end{align*}
which leads to the Lagrangian function described as:
\begin{equation}
	L = K - V = \frac{m}{2} l^2 \left(\dot{\theta}^2 + s_\theta^2 \dot{\phi}^2\right) + \frac{m}{2}\dot{l}^2 + mgl c_\theta
\end{equation}
The dynamics of the system will be obtained using the Euler-Lagrange equation:

\begin{equation}
\begin{aligned}
		&\ddot{\theta} + \frac{2}{l} \dot{\theta} \dot{l} - c_\theta s_\theta  \dot{\phi}^2 + \frac{g}{l}  s_\theta =  \frac{1}{ml}F_{u,n} \\
		&\ddot{\phi} + 2 \frac{c_\theta}{s_\theta} \dot{\theta} \dot{\phi} + \frac{2}{l} \dot{\phi} \dot{l} = \frac{1}{ml s_\theta} F_{u,t} \\
		&\ddot{l} - l \dot{\theta}^2 - l s^2_\theta \dot{\phi}^2 - g c_\theta = \frac{1}{m} F_r .
\end{aligned}
\label{eq:nonlinearDyn}
\end{equation}

Clearly, in this derivation of the model, we have heavily
relied on the point-mass nature of the body.  A possible issue
could be the rotational dynamics of the body around the rope axis,
which could waste energy, generating unnecessary motions and
impede the landing phase. This issue will be part of our
future work. However, as discussed next, under reasonable
assumptions (e.g., thrusting force oriented in the direction
of the center of mass), the results based on the simplified
model can be applied to a realistic system with a good approximation.
\vspace{-0.2cm}
\subsection{Robot motion: problems and solution overview}
The problem of moving CLIO between two given configurations is not 
easy.  First of all, even the simplified dynamics in~\eqref{eq:nonlinearDyn} is highly nonlinear.  Second, 
the configurations  between which  the robot moves are usually 
distant, and therefore, using a linearization of the dynamics equations \ref{eq:nonlinearDyn} 
around a specific operating point would lead to 
inaccurate results.  Third, one of our
actuators, the thrusting force $\mathbf{F}_u$, has an impulsive nature
and operates at discrete time instants. Therefore, it is not possible
to use it in any feedback control scheme.
Finally, the tangential component $F_{u,t}$ is generated by
using friction. Therefore, the two components $F_{u,n}$ and $F_{u,t}$ are coupled
by the nonlinear friction cone constraint. Because the friction cone
depends on the specific point or area where the robot is pointing its leg,
this mechanism is not totally reliable (i.e., there can be significant deviations
between generated and desired values of $F_{u,t}$).

In view of this complexity, we propose an approach based on two steps:
1. a motion strategy is planned prior to starting the motion,
2. after take off, a feedback motion controller operates on
$\mathbf{F}_r$ to compensate for small deviations and
secure that the robot lands close to the expected position.
In this paper we will mainly focus on step 1.

\section{Motion Planning}\label{sec:motionP}
The motion plan for the robot is decided by solving an optimal control problem \emph{\`a la} Pontryagin.
Generally speaking, we can set up the problem in the following way: \vspace{-0.5cm}

\begin{equation}\label{eq:OCP}
  \begin{aligned}
    & J = \min_{\vect{u}(t)} \,M(\vect{x}(t_0),\vect{x}(t_f),t_0,t_f)+\int_{t_0}^{t_f}   L\left(\vect{x}(t),\vect{u}(t)\right) dt\\
    &\text{subject to:}\\
    &\,\,\dot{\vect{x}}(t) = \vect{f}(\vect{x}(t),\vect{u}(t)),\\
    &\,\,\vect{u} \in \mathcal{U},\\
    &\,\,\vect{h}\left(\vect{x}(t),\vect{u}(t)\right) \leq 0,\\
    &\,\,\vect{B}(\vect{x}(t_0), t_0, \vect{x}(t_f), t_f)=0,
  \end{aligned}
\end{equation}

where $\vect{x}$ is the vector of state variables, $\vect{u}$ is the control input constrained in the convex set $\mathcal{U}$, $f$ is the dynamic equation of the system,
$\vect{h}$ is the constraint on the state and control variables, $\vect{B}$ are the boundary conditions, $t_0$ and $t_f$ are the initial and final time (that is an optimization variable), which can be part of the optimization. The OCP~\eqref{eq:OCP} is in the standard form with Mayer term $M$ and Lagrange term $L$ that model initial/terminal costs and  the running cost, respectively.\\
\noindent
{\bf Dynamical System.}
The state $\vect{x}$ of our problem is $\vect{x} = \left[\theta,\,\phi,\,l,\,\dot{\theta},\,\dot{\phi},\, \dot{l}\right]^T$, while the control variables are
given by $\vect{u} = \left[\vect{F}_u,\,\vect{F}_t\right]^T$. The dynamic equation is derived from~\eqref{eq:nonlinearDyn}:
\begin{equation}
  \label{eq:NonLinStateSpace}
  \begin{aligned}
      \begin{bmatrix}
        \dot{x}_1\\\dot{x}_2 \\ \dot{x}_3\\ \dot{x}_4\\ \dot{x}_5\\ \dot{x}_6\end{bmatrix}
     =
    \begin{bmatrix}
      x_4\\
       x_5\\ x_6 \\  - \frac{2}{x_3} x_4 x_6 + c_{x_1} s_{x_1}  x_5^2 - \frac{g}{x_3}  s_{x_1} +  \frac{1}{mx_3}F_{u,n}\\
       - 2 \frac{c_{x_1}}{s_{x_1}} x_4 x_5 - \frac{2}{x_3} x_5 x_6 + \frac{1}{mx_3 s_{x_1}} F_{u,t}\\
       x_3 x_4^2 + x_3 s^2_{x_1}x_5^2+g c_{x_1} +\frac{1}{m}F_r .
     \end{bmatrix}
    \end{aligned}
\end{equation}

\noindent
{\bf Boundary Conditions.}
The terminal constraints are usually expressed in Cartesian space $[X,\,Y,\,Z,\,\dot{X},\dot{Y},\dot{Z}]^T$ and they can be expressed as a function of
the state variables by means of~\eqref{eq1}. For instance, for $X,Y,Z$, we have:
\begin{equation}
  \begin{aligned}
    &x_1= \atandue\left(-Z,\sqrt{X^2+Y^2}\right), \\ 
    &x_2 = \atandue(Y,X),\\
    &x_3 = \sqrt{X^2+Y^2+Z^2},\quad \\
  %  &x_4(t_0) = \frac{1}{\Lambda}\left(c_{x_{1}(t_0)} c_{x_2(t_0)} \dot{X}_0 +c_{x_1(t_0)}s_{x_2(t_0)} \dot{Y}_0+s_{x_1(t_0)}\dot{Z}_0\right)\\
 %   &x_5(t_0) = \frac{1}{\Lambda s_{x_1(t_0)}}\left(-s_{x_2(t_0)} \dot{X}_0) \right) + \\
 %   &\quad  +  \frac{1}{\Lambda s_{x_1(t_0)}} \left(c_{x_2(t_0)} - c_{x_2(t_0)} c^2_{x_1(t_0)} + c^2_{x_1(t_0)} s_{x_2(t_0)}  \right) \dot{Y}_0 \\
%    &\quad + \frac{1}{2 \Lambda}\left(-c_{x_1(t_0)}\left(\sqrt{2}\sin\left(2 x_2(t_0)+\pi/4\right)-1 \right) \right) \dot{Z}_0 \\
%    &x_6(t_0) = \frac{x_3(0)}{\Lambda}\left(s_{x_{1}(t_0)} c_{x_2(t_0)} \dot{X}_0 + s_{x_1(t_0)}s_{x_2(t_0)} \dot{Y}_0\right)\\
%    & \quad + \frac{1}{2 \Lambda} c_{x_1(t_0)}\left(\sqrt{2} \cos\left(2 x_2(t_0)+\pi/4\right)-\frac{1}{2}\right) \dot{Z}_0 \\
 %   &\Lambda = x_3(t_0) \left(c_{x_2(t_0)} s_{x_2(t_0)} c^2_{x_1(t_0)} - c^2_{x_2(t_0)} c^2_{x_1(t_0)} + 1\right).
  \end{aligned}
\end{equation}

\noindent
{\bf State Constraints.}
State constraints are related to regions of the operation space that are not accessible. For instance, an irregular form of the walls
could generate obstacles that the robot 
has to overcome in order to reach its final destination. Generally speaking, the inaccessible area is modelled as
a region whose boundary is a surface. We assume that this surface can be expressed by
a differential 2D manifold expressed by $f(\vect{x}) = 0$. Therefore the admissible region is generally given by
$f(\vect{x}) \geq 0$ and could be non-convex. 
%By using~\eqref{eq1}, we can express the constraint either in Cartesian coordinates or in terms of the state variables. 
Another constraint is added to prevent collision with the wall (i.e. $X >0$) and to ensure the robot elevation does not exceed the anchor level (i.e. $Z<0$).

\noindent
{\bf Actuation Constraints.}
The system actuators are given by the two forces $\mathbf{F}_r$ and $\mathbf{F}_u$.
The $\mathbf{F}_r$ force acts along the direction of the rope ($\vect{e}_R$ axis). % \vect{f}_r$ (see~\eqref{eq:forces}) this is for journal
Since the rope cannot push the robot 
(assumption 3, the rewinder can only pull the wire), $F_r$ can only be used to retract the robot or to slow down its descent (under the action of gravity). 
Moreover, the force is bounded by the limits of the actuators (e.g. a hoist). Therefore, the
constraint on $F_r$ is:
\[
  -F_r^{\text{max}} \leq F_r \leq 0.
\]
As regards $\mathbf{F}_u$, since it is meant to act for a very short duration, reaching high peaks, we model it through a Dirac impulse: $\mathbf{F}_u   (t) = \mathbf{F}_u \delta(t)$, where
$\mathbf{F}_u$ is a vector with the two components $F_{u,t}$ and $F_{u,n}$.
The Dirac delta is a generalized function and cannot be handled by the optimal control frameworks, therefore, a smooth approximation is needed.
A natural approximation is a Gaussian function, that is $\delta(t-t_0) \approx \frac{1}{\sqrt{2 \pi \sigma^2}} e^{-\frac{(t-t_0)^2}{2 \sigma^2}}$, \cite{saichev:1996}. The duration $T_{th}$ of the impulse 
  is roughly given by $6 \sigma$. Given the nature of our actuation mechanism, a realistic choice is to have a duration in the order of tens of milliseconds, which consequently
  determines a remarkable intensity of the impulse. The maximum norm of $\mathbf{F}_u$ is upper-bounded by the actuation limits:
  \[
    \sqrt{F_{u,n}^2 + F_{u,t}^2} \leq F_{u}^{\text{max}} \quad .
  \]
  Additionally, the tangential component $F_{u,t}$ is generated by the friction with the mountain wall. 
  Hence, $F_{u,n}$ and $F_{u,t}$ are constrained by the following relation (friction cone):
\begin{align*}
\vert F_{u,t} \vert \leq \mu F_{u,n}
\end{align*} 
where $\mu$ is the \textit{friction coefficient}, a constant depending on the nature of the terrain. 
%For our simulation scenarios below, we have chosen $\mu = 0.8$ and $F_u^{\text{max}}=1000$ N. 

\noindent
{\bf Objective Function.}
%The cost functional is composed of two terms: an integral component accounting for the cost accumulated during a period of time, and a terminal component, which accounts for the cost related to the final configuration of the system.
The cost is composed of three terms. 
First, we minimize the time $t_f$ to complete the mission (assuming without loss of generality that $t_0=0$). 
Second, we minimize the final kinetic energy $K(t_f)$ because when the robot lands, it has to dissipate the energy through a damper (we are not considering to reuse the accumulated energy to re-bounce after landing).
Third, we minimize the work performed by the winding mechanism.
Thus, the cost is:
\begin{align*}
&J = w_tt_f + w_E K(t_f) + w_{F_r} \int_{0}^{t_f} (F_r\cdot x_6(t))^2 dt ,  \\
& K(t) = \frac{m}{2} x_3(t)^2 \left(x_4(t)^2 + s_{x_1(t)}^2+ x_5(t)^2 \right) + \frac{m }{2} x_6(t)^2,
\end{align*}
%%%
with $w_t$, $w_E$ and $w_{F_r}$ being the weights associated to the three cost components.
%We are interested in minimizing the interval of time $t_f - t_0$ needed to complete the mission and the kinetic energy $K(t_f)$ at the final instant. The reason for $K(t_f)$ is because when the robot lands,
%it has to stop its motion and dissipate the energy through a damper (at the moment we are not considering to reuse the accumulated energy to re-bounce after landing).
%Therefore, assuming without loss of generality that $t_0=0$, we optimize
%\begin{align*}
%&J = w_tt_f + w_E K(t_f) + w_{F_r} \int_{0}^{t_f} (F_r\cdot x_6(t))^2 dt ,  \\
%& K(t) = \frac{m}{2} x_3(t)^2 \left(x_4(t)^2 + s_{x_1(t)}^2+ x_5(t)^2 \right) + \frac{m }{2} x_6(t)^2,
%\end{align*}
%%%%
%with $w_t$ and $w_E$ two weights to combine the temporal with the energetic contribution while $w_{F_r}$ is the weight 
%associated to a term proportional to the work performed by the winding mechanism.
We should observe that, in case of a difficult convergence, it is possible to ``soften'' some of the constraints. For instance, for the terminal constraint, it is possible to introduce a
slack variable $\Delta$ and have a constraint $\left\|\vect{p}(t_f) - \vect{p}_f\right\|^2 \leq \Delta$, where the slack $\Delta$ can be either a maximum tolerance set by the user or become part of the cost function. Another possibility is to add the relaxed constraint as a penalty in the objective function.

\noindent
\textbf{Initial guess.}
To speed-up convergence, we initialized the optimization with a reasonable initial guess.
Because time $t_f$ is an optimization parameter, we compute the time constant for the system linearized around the initial state. %The other variables were initialised ...
The linearized system becomes also a good approximation when the jump length is small with respect to the rope elongation.

%\section{Motion Control} \label{sec:motionC}
%\input{sections/control}

\section{Simulation Results}\label{sec:results}
This section presents simulation results that show how the control inputs
obtained running the optimization of Section \ref{sec:motionP}, are able to bring the robot to a desired target. 
We used the OCP solver PINS~\cite{pins:1,pins:2,pins:3} which relies on the indirect approach based on the Pontryagin Maximum Principle.
The code to replicate the results is freely available at \footnote{ \href{https://github.com/mfocchi/climbing\_robots.git}{https://github.com/mfocchi/climbing\_robots.git}}.

% 1st experiment: validation of optimization with 2 models, with obstacle
In a first experiment, we run the optimization to perform  a 12 $m$ long jump starting from an initial 
position $\vect{p}_0 = \mat{0.24& 0& -8}^T$ up to a target at $\vect{p}_f = \mat{3 & 3 &-20}^T$ $[m]$. As an additional constraint,
during the jump, the robot has to avoid a rock pillar attached on the wall, that we model as a 20 $m$ high cone with a base of radius 2.5 $m$.
We validate the results of the optimization for both the simplified model (in a Matlab environment)
and the full 3D model. For this case, we built a Gazebo simulator based on the URDF description \cite{urdf} of the robot (see Fig. \ref{fig:3dmodel}(right)).
Table \ref{tab:params} reports the used physical parameters, together with some optimization settings.
\begin{table}[!tbp]
	\centering
	\caption{ Simulations parameters}
		\begin{tabular}{l c c  } \hline\hline
			\textbf{Name} \quad                  & \textbf{Symbol}                     & \textbf{Value}  \\ \hline
					Robot mass                   & m  & 5                 \\ 
					Max. impulse  [N]       	 & $F_{u}^{\text{max}}$				& 1000   \\
					Max. retraction force [N]    & $F_r^{\text{max}} $			& 200    \\
					Friction coeff.				 & $ \mu $ 					        & 0.7   \\
					Thrust impulse duration	[s]  & 	$T_{th}$  				& 0.025\\
					Discretization steps         & N 						& 500\\					        
			\hline\hline 					    					    					    
		\end{tabular}
		\vspace{-0.5cm}
		\label{tab:params}
\end{table}
For the Matlab simulation, we simply integrate~\eqref{eq:nonlinearDyn} 
with an ode45 Runge-Kutta variable-step scheme.
For the Gazebo simulation, we assume to apply the push impulse as a force $F_c$ at the contact point.
Therefore we employ the leg dynamics to define the mapping between the contact force $F_c$ and the torques $\tau_{\text{leg}}$ at the leg joints:
\begin{equation}
\boldsymbol{\tau}_{\text{leg}}^d= \mat{\tau_{HP} \\ \tau_{HR} \\ \tau_{K}} =  \vect{h}_{\text{leg}} -\vect{J}_{\text{leg}}^T \underbrace{\left(R_e^\mathcal{W}  \mat{F_{u,n} \\ F_{u,t} \\ 0} \right) 	}_{\vect{F}_c}
\end{equation}
Where $\vect{J}_{\text{leg}} \in \Rnum^{3 \times 3}$ is the sub-matrix of the Jacobian $\vect{J}$ relative to the leg joints, 
%$R_e^\mathcal{W}$ is the rotation matrix that represents the orientation of the base link w.r.t. the inertial frame $\mathcal{W}$ (different from $R_e^\mathcal{W}$) 
and $\vect{h}_{\text{leg}} \in \Rnum ^3$ represents the bias terms. We do
not generate any impulse along the rope direction ($\vect{e}_R$ axis) in 
order to avoid to accidentally create any slack on the rope.
Additionally, to avoid angular motions, we implement virtual dampers to keep the passive joints at the 
base in a fixed position, since the optimization is performed 
considering the simplified model that neglects the angular dynamics. 

\begin{figure}[!tbp]
	\centering
	\includegraphics[width=0.8\columnwidth]{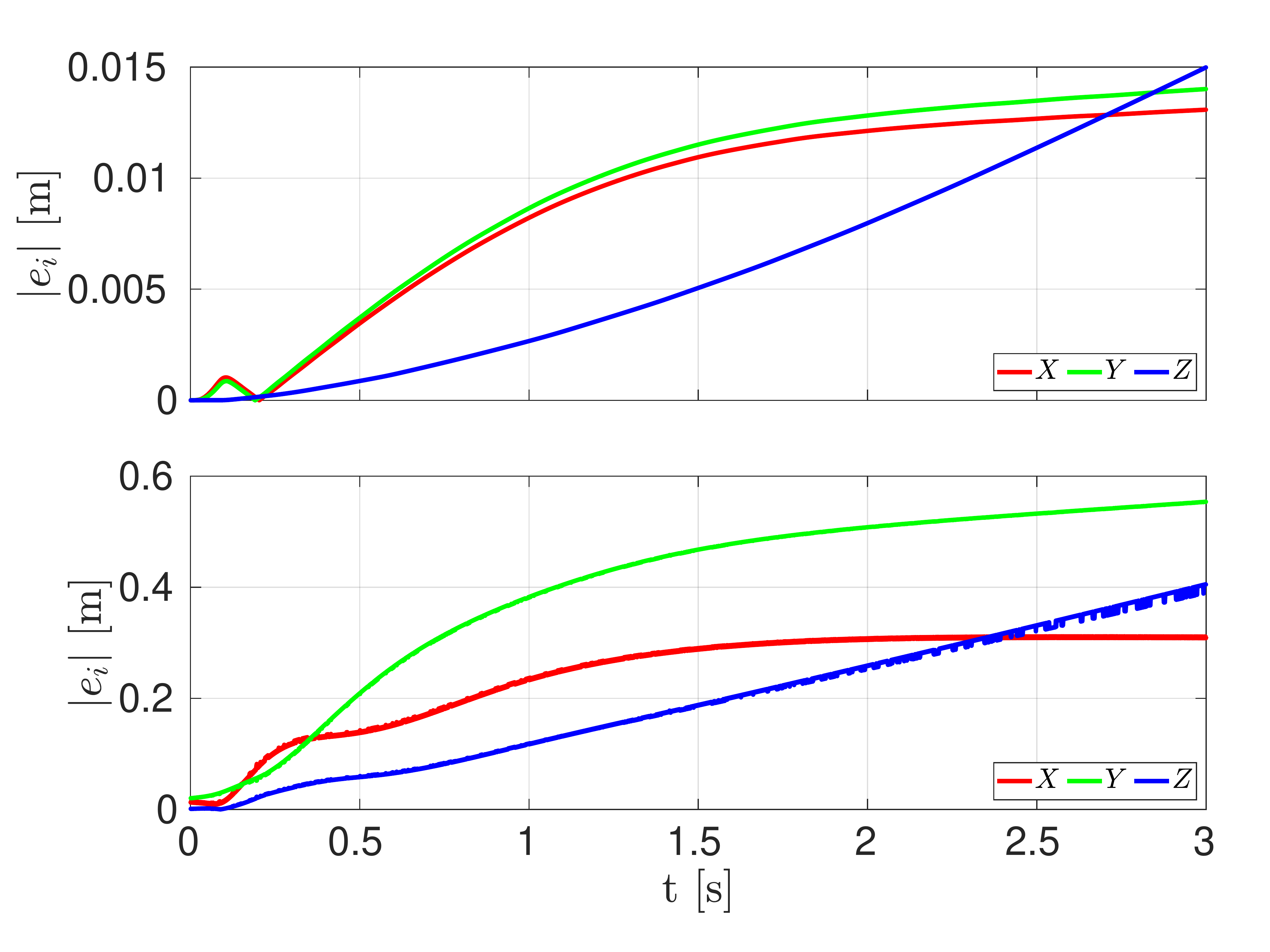}
	\caption{\small Simulation. Validation of the optimization results.  Absolute value of the validation error between 
		the \gls{com} trajectory computed by the optimization and the simulated trajectory with (upper plot) Matlab and (lower plot) Gazebo, respectively. 
		%The bottom plot is the rope retraction force $F_r(t)$.
		}
	\label{fig:validation}
		\vspace{-0.65cm}
\end{figure}

We set the initial configuration to $\vect{q}_0= \mat{ \atandue(r_{\text{leg}}, l_0) & 0 &0  & l_0 & 0 &  0 & 0  & -1.57  & 0 & 0}^T$ 
where $l_0$ is the initial rope length and $r_{leg}=0.32$ $m$ is the leg length at the startup configuration. 
At $q_0$,  the foot is meant to touch the wall in $\vect{p}_0$, the starting point of the optimized trajectory.
A state machine coordinates the 3 phases of the jump: leg orientation, thrusting and flying. 
In the leg orientation phase,  the hip roll joint set-point $q_{HR}^d$ is commanded
to have the leg aligned with the impulse direction, $q_{HR}^d = \atandue(\text{max}(F_{u,t}), \text{max}(F_{u,n}))$. 
Note that the hip-pitch joint set-point is $q_{0, HP}^d = -1.57$ in order to 
have the leg lying on the X-Y plane of the base frame. A low level PD controller 
runs in parallel with the  feed-forward actions $\boldsymbol{\tau}_{\text{leg}}^d$ to drive the joints.  
During the \textit{thrusting} phase, the force $\vect{F}_c$ is generated at the contact 
applying $\boldsymbol{\tau}_{\text{leg}}^d$ for the thrust duration $T_{th} = 0.025 s$, while the PD gains are 
switched off to avoid conflicts. Then  it follows  the 
flying phase, where   the rope winding joint is actuated with the optimized force pattern $\tau_R = F_r(t)$ for the whole jump duration $t_f$.
%
% computation time and number of nodes
The computation time for the optimization and the integration error at the target $\Vert e_f \Vert$ are linearly  
linked to the number of discretization points $N$. Table~\ref{tab:solve_time} reports 
the integration error normalized for the jump length $\Vert \vect{e}_f \Vert / (l_f-l_0)$ 
for different numbers of discretization points. 
%%%
\begin{table}[!tbp]
\centering
	\caption{ Results of the numerical OCP for different discretisations}
	\label{tab:solve_time}
	\begin{tabular}{c c c c  } \hline\hline
		\textbf{N} \quad & \textbf{Comp. time [s]}           & \textbf{$\Vert \vect{e}_f \Vert$ [m]} &  \textbf{$\Vert \vect{e}_f \Vert / (l_f-l_0)$ [\%]} \\ \hline
			250    &   0.850          &  0.0937 & 0.76 \\
			500    &   1.472          &  0.0525 & 0.42 \\
			1000   &   2.648          &  0.0238 & 0.19 \\
			2000   &   5.063          &  0.0180 & 0.14 \\
		\hline\hline 					    
	\end{tabular}	
		\vspace{-0.4cm}
\end{table}
The Table shows that a  good trade-off can be found between accuracy and computation time that enables fast computation.
Fig. \ref{fig:validation} reports the results for $N=500$. 
Simulating the simplified model, the matching is almost perfect (apart from integration drifts), while in the 
full-model Gazebo simulation, the norm of the final error is around 0.75 $m$ for a 16 $m$ jump ($<5\%$). This is due to a number of reasons: 
1) the control inputs are applied in open-loop, therefore any uncertainty causes an error in the 
lift-off momentum that can cause drift during the jump, 2) the impulse is applied at the foot and not at the \gls{com}, 3) the approximation due to  
the simplified model with respect to the full one used in the Gazebo simulation.
However, a 0.6\% tracking error is in a range that can be efficiently coped with a controller, 
and shows that the simplified model is a good approximation for the real system. 

\begin{figure}[!tbp]
	\centering
	\includegraphics[width=0.7\columnwidth]{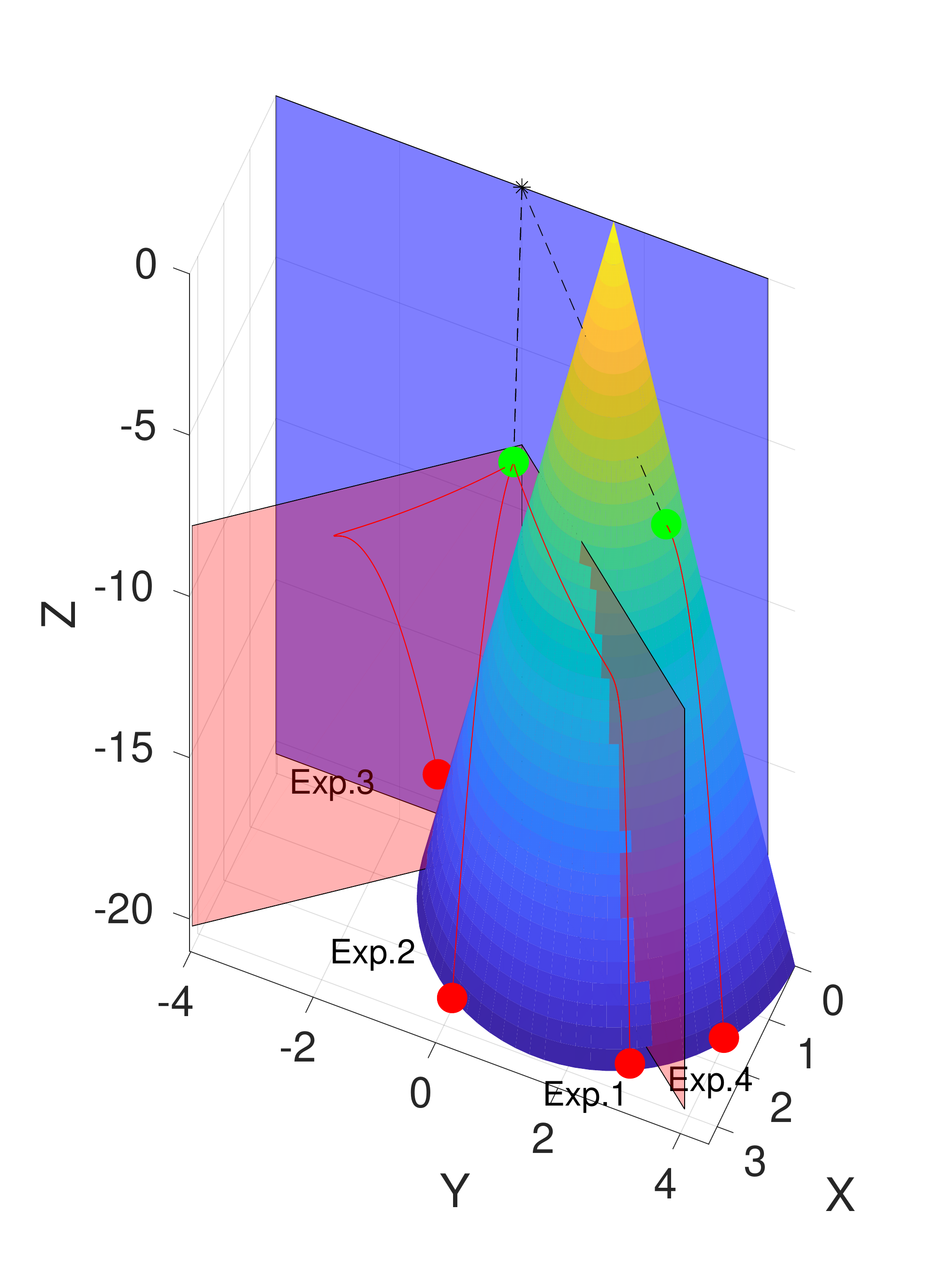}
			\vspace{-0.5cm}
	\caption{\small Simulation. The black star is the anchor point, the green dot is the starting point $\vect{p}_0$ the red dot the target location $\vect{p}_f$.
			The blue shaded area is the rock face,  while the red	shaded  area  represents 
			the friction boundaries ($\mu = 0.7$) in the tangential direction. The  cone represents an obstacle (rock pillar) to be avoided.}
		\vspace{-0.5cm}
	\label{fig:targets}
\end{figure}

%
%
% 2nd experiment multiple targets, an one starting from slanted
Additionally, we run the optimization to jump on 3 different targets on the rock pillar 
(see Fig. \ref{fig:targets}), starting from a point on the wall $\vect{p}_0 = \mat{0.24,0,-8}$ (Experiment 1,2,3).
To demonstrate that the approach is valid also for  jumps from a \textit{non vertical} (i.e. slanted ) surface, 
Experiment 4 is a jump from  a location $\vect{p}_0=\mat{0.63 & 2.35& -7.5}$ that is already on the pillar, which has an inclination of around 80 $deg$.
Each optimization provided the max values of the initial impulses ($F_{u,n}$, $F_{u,t}$)\footnote{The value of these forces are related to the selected impulse duration $T_{th}$  [s], they can be strongly reduced by taking longer durations (i.e. in accordance to the actuator  response time).}, the pattern of the winding force $F_r$ and the jump duration $T_f$.
For the first 3 targets, we also plot the  friction boundaries in the tangential direction at the lift-off point (red shaded  area).

\begin{figure}[!tbp]
		\centering
	\includegraphics[width=0.8\columnwidth]{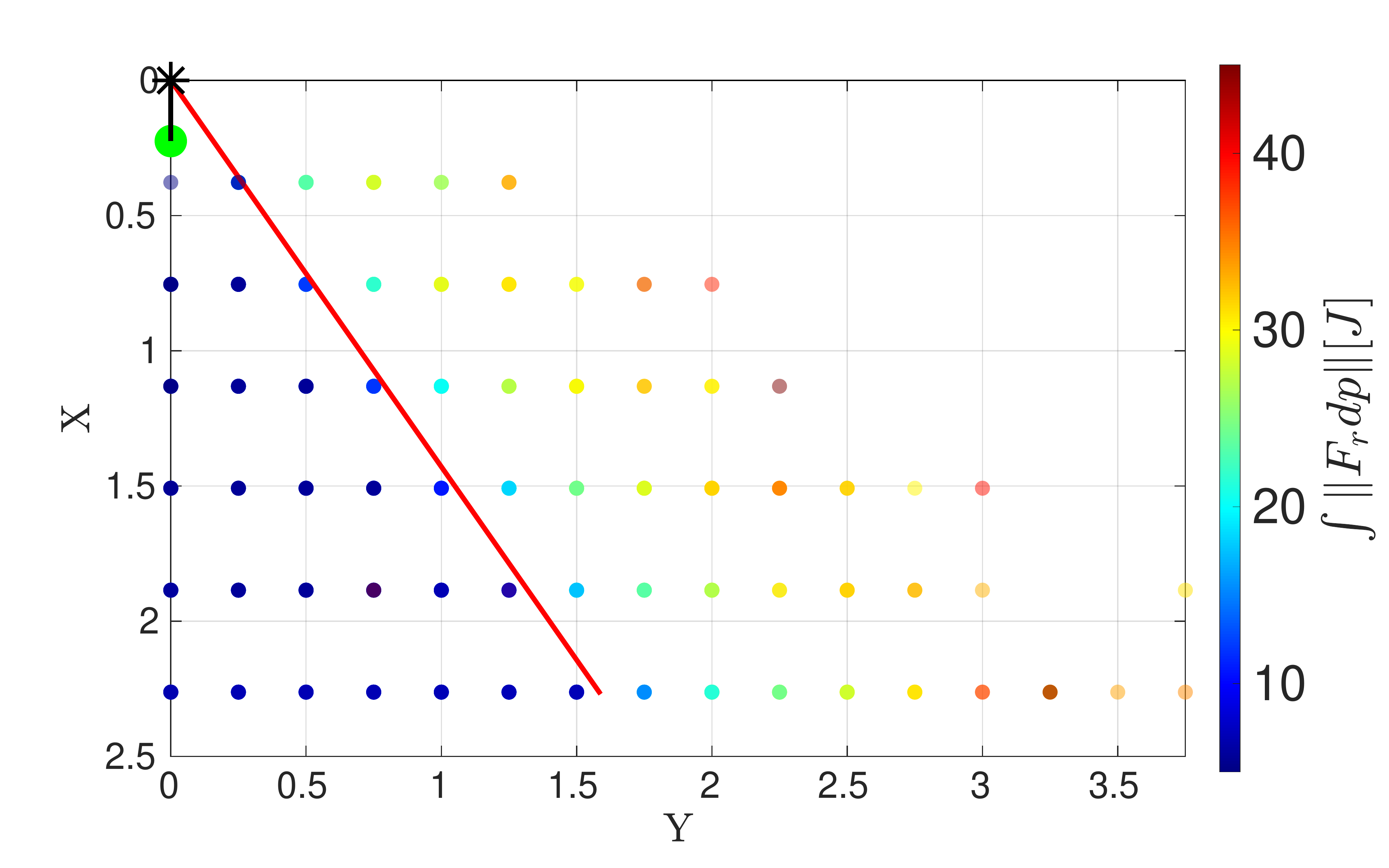}
	\caption{\small Simulation. Top view (X-Y plane) of the region of reachable targets for a friction coefficient $\mu =0.7$.
		The black star is the anchor point, the big green dot is the common lift-off point $\vect{p}_0$. The color bar indicates the energy consumed by the winding mechanism to reach the target.}
	\label{fig:reachable_region}
			\vspace{-0.5cm}
\end{figure}

The results are reported in Table \ref{tab:sim_different_targets} 
and in the accompanying video\footnote{Video link: \href{https://youtu.be/RZUfaaDkoR0} 
{https://youtu.be/RZUfaaDkoR0} }. 
As expected, the final kinetic energy (that will be lost at the impact) is higher for longer jumps, the jump duration, instead, is higher for the jumps that involve a bigger lateral motion.
\begin{table}[!tbp]
	\centering
	\caption{Multiple Targets Simulations Results}
	\resizebox{\columnwidth}{!}{
		\begin{tabular}{c   l c c c c  c } \hline\hline
			\textbf{Exp.}   & \textbf{$\vect{p}_f$ [m]}           & \textbf{$T_f$ [s]} &  \textbf{$F_{u,n}$ [N]} & \textbf{$F_{u,t}$ [N]} &\textbf{ $K_f$ [J]}\\ \hline % & \textbf{$\Vert e \Vert$ [m]}
			1               & $\mat{2.60, 2.73, -20.46} $    &      3.12       &    306                 &      214           &  578       \\ % exp2
			2               & $\mat{ 2.50,  0.05,   -20.46}$ &      1.62       &     236                 &       5.11           &   596                \\ % exp3
			3               & $\mat{0.55, -0.8,  -17.76} $   &     3.18         &   217                       & -151         &        422        \\ % exp5
			4   			& $\mat{1.73, 3.8,  -20.46} $     &     1.69         &       168                &     -64          &      619     &                 \\ % exp6_slanted		
			\hline\hline 					    					    					    
	\end{tabular}}
		\vspace{-0.4cm}
	\label{tab:sim_different_targets}
\end{table}
By inspecting the red trajectories in Fig. \ref{fig:validation}, one can see that for the targets close to the friction boundaries, 
to "steer" the trajectory more laterally, the optimizer  dictates  an \textit{initial} 
retraction of the rope. This strategy is meaningful, because  lateral pushes are limited due to the limits on the tangential force posed by friction, 
therefore the only resort is  to vary the time constant of the "pendulum" (i.e.  reducing the  inertia moment about the anchor point) by initially winding the rope.
This  decreases the deceleration on the $\phi$ variable due to the gravity component. Eventually, the rope is let to passively unwind  (under the action of gravity) to  attain the target. 
%Journal
%An interesting outcome of this analysis is that, despite the system is fully controllable (locally), 
%the reachable targets, when jumping  from the anchor line, are limited to the area inside the cone, 
%therefore to move tangentially \textit{away} from the anchor line  
%only small jumps can be performed. Conversely, the jumps have no limit when jumping toward the anchor line, 
%because the gravity component (always pointing toward the anchor line) can be exploited. 
%
%
To  visualize this fact and better assess the capabilities of the system, we perform a reachability analysis, plotting a top view (X-Y plane) of the region  of reachable jump targets (see Fig. \ref{fig:reachable_region}) lifting-off from the same point $\vect{p}_0$.  We run a number of optimizations for a grid of targets below $\vect{p}_0$, in the positive $Y$ half-space (the solutions are symmetric in the negative one). We plot a marker for the targets where the optimization convergence is successful to state that they are \textit{reachable}. We also  associate a color code to the energy consumed by the winding mechanism for each target. The friction boundary is marked as a red line. We notice that there are a few reachable targets  out of the area limited by the friction cone and that they  are associated with a very high consumption in terms of energy from the winding mechanism, because they involve a big initial retraction, corroborating the insights got in Fig. \ref{fig:targets}. On the other hand, with a fixed rope,  lateral jumps would be very limited. Indeed, the fact of performing small lateral jumps to move laterally, is also what experienced climbers do to realize a big pendulum swing on a rock face \cite{kingswing}.
%We noticed that, due to the physics of the system, in order to reach lateral targets the optimization tried to rewind the rope 
%to steer the trajectory that would be otherwise limited to an area linked to the value of the friction coefficient.
%We selected this quite large value for $\mu$ because it is reasonable to assume that the 
%robot will exploit the asperities present on a rocky wall to perform the push, which provide a reasonable amount of tangential impulse. 

\section{Conclusions}
\label{sec:conclusion}
We presented a  robotic solution for exploration and rescue 
missions in mountain environments such as canyons, lunar craters, etc. 
It could be employed in civil applications in scaling maintenance activity, e.g., to detach from the
mountain walls dangerous boulders,  needed to mitigate the hydro-geological risk.
%or loose and potentially unstable vegetation (a.k.a. scaling) or to apply landslide protection networks
%maintenance costs in scaling application
This would create additional market opportunities for a possible business.
Our platform combines the use of a rope of adjustable length with a jumping mechanism.
We validated the result of the optimization carried out using a simplified model 
with a fully-detailed model  simulation in Gazebo,
obtaining  $5\%$ discrepancy on a 16 $m$ jump.
% limitations
The main limitations of the actual work are the limited \textit{reachability} with 
the single anchor arrangement and the lack of descriptiveness (in terms of angular dynamics) of the simplified model.
This prevents to devise planning strategies that optimize also the orientation of the robot at the landing. 
To address the first issue, we plan to investigate a solution with two ropes with more than one anchor point to increase the reachable region on the wall;
while, for the second issue, we plan to release the point-mass approximation and considering the robot as a rigid body with
non-trivial mass geometry. In this case, the thrust action has to be combined with a way to control the rotation of the body 
and in order to ensure the proper alignment of the leg  to guarantee a safe landing.  
Last, we are working on the design of a landing controller to dissipate the excess of kinetic energy at landing, 
avoiding rebounces. All these steps are preparatory to the development of a working prototype.
Many important future directions have been opened by this research.
Whilst avoiding obstacles is a well-known problem for motion planning in horizontal terrain, 
the jump motion pattern determined by the ropes makes the motion-planning
problem non-standard, calling for new approaches.
Much work has to be done also in the area of motion control in order to
ensure that the planned trajectory is closely tracked during the flight.

\small
\bibliographystyle{IEEEtran}
\bibliography{references/references}

%\section{Acknowledgements}
%The publication was created with the co-financing of the European Union FSE-REACT-EU, PON Research and Innovation 2014-2020 DM1062 / 2021.

\end{document}